\documentclass[11pt,a4paper]{article}
\usepackage[hyperref]{naaclhlt2019}
\usepackage{times}
\usepackage{latexsym}

\usepackage{url}
\usepackage{subfig}
\usepackage{graphicx}

\aclfinalcopy %

\usepackage{latexsym}
\usepackage{multirow}
\usepackage{arydshln} 
\usepackage{enumitem}
\usepackage{booktabs}
\usepackage{newtxtext,newtxmath}

\usepackage{xparse}   %
\NewDocumentCommand{\rot}{O{45} O{1em} m}{\makebox[#2][l]{\rotatebox{#1}{#3}}}%

\usepackage{amsmath}
\usepackage{amssymb}
\DeclareMathOperator*{\argmax}{arg\,max}

\newcommand{\cut}[1]{}

\newcommand{\bp}[1]{\textcolor{magenta}{\textbf{BP: #1}}}

\title{Beyond task success:\\ A closer look at jointly learning to see, ask, and GuessWhat}
\author{Ravi Shekhar$^\dagger$, Aashish Venkatesh$^*$, Tim Baumg\"{a}rtner$^*$, Elia Bruni$^*$, \\ 
\textbf{ Barbara Plank$^\varheartsuit$, Raffaella Bernardi$^\dagger$ \and Raquel Fern\'{a}ndez$^*$} \\
$^\dagger$University of Trento, $^*$University of Amsterdam,$^\varheartsuit$IT University of Copenhagen  \\
    {\tt bplank@itu.dk} \ \  {\tt raffaella.bernardi@unitn.it} \ \ {\tt raquel.fernandez@uva.nl} }

\date{}

\begin{document}
\maketitle
\begin{abstract}
We propose a grounded dialogue state encoder which addresses
a foundational issue on how to integrate visual grounding with
dialogue system components. As a test-bed, we focus on the
\textit{GuessWhat?!}~game, a two-player game where the goal
is to identify an object in a complex visual scene
by asking a sequence of yes/no questions. Our visually-grounded
encoder leverages synergies between guessing and asking questions,
as it is trained jointly using multi-task learning.
We further enrich our model via a cooperative learning regime.
We show that the introduction of both the joint architecture and cooperative   
learning lead to accuracy improvements over the baseline system.
We compare our approach to an alternative
system which extends the baseline with reinforcement learning. Our
in-depth analysis shows that the linguistic skills of the
two models differ dramatically, despite approaching comparable   
performance levels. This points at the importance of analyzing
the linguistic output of competing systems beyond numeric comparison
solely based on task success.\footnote{Equal contribution by R.~Shekhar and A.~Venkatesh.}
\end{abstract}

\section{Introduction}
\label{sec:introduction}

Over the last few decades, substantial progress has been made in developing dialogue
systems that address the abilities that need to be
put to work during conversations: Understanding and
generating natural language, planning actions, and tracking the 
information exchanged by the dialogue participants. The latter is particularly critical since, for communication to be
effective, participants need to represent the state of the dialogue and the common ground
established through the conversation \cite{Stalnaker1978,Lewis1979,clark1996using}.

\begin{figure}[t]\centering \hspace*{-0.1cm}
\includegraphics[width=\columnwidth]{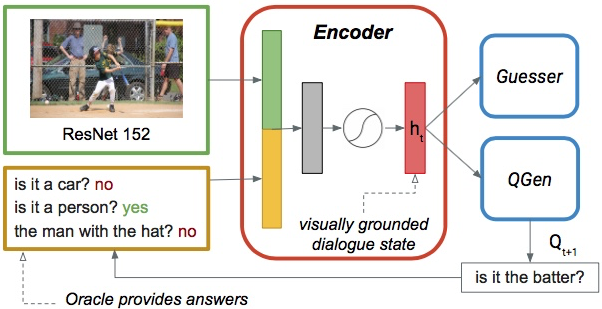}
\caption{Our questioner model with a single visually grounded
  dialogue state encoder.}\label{fig:model-intro}
  \vspace*{-10pt}
\end{figure}

In addition to the challenges above, dialogue is often situated in a perceptual environment. 
In this study, we develop a dialogue agent that builds a 
representation of the context and the dialogue state 
by integrating information from both the \emph{visual} and \emph{linguistic} modalities.
We take the {\em GuessWhat?!}~game \cite{guesswhat_game} as our test-bed, a two-player game
where a Questioner faces the task of identifying a target object in a visual scene by asking a series of yes/no questions to an Oracle.
We model the agent in the Questioner's role. 

To model the Questioner, previous work relies on two independent models to learn 
to ask questions and  to guess the target object, each equipped with its own encoder~\cite{guesswhat_game,stru:end17,zhu:inte17,lee:answ17,shek:askn18,zhan:aski}.
We propose an end-to-end architecture with a single \emph{visually-grounded dialogue state encoder} (cf.\ Figure~\ref{fig:model-intro}). %
Our system is trained jointly in a supervised learning setup, extended
with a cooperative learning (CL) regime: By letting the model play the
game with self-generated dialogues, the components of the Questioner agent 
learn to better perform the overall Questioner's task in a cooperative manner.
\citeauthor{visdial_rl}~\shortcite{visdial_rl} have explored the use
of CL to train two visual dialogue agents  that receive joint rewards when they play a game successfully. To our knowledge, ours is the first approach where cooperative learning 
 is applied to the internal components of a grounded conversational
 agent. 
 
 Our cooperative learning regime can be seen as an interesting alternative to 
 reinforcement learning (RL)---which was first applied to {\em GuessWhat?!} 
 by~\citet{stru:end17}---because it is entirely differentiable and computationally 
less expensive to train than RL.
Little is known on how this learning approach compares to RL
 not only regarding task success,
 but also in terms of the quality of the linguistic output, a gap we seek to fill in this paper. 
In particular, our contributions are:\footnote{Code and supplementary material are available at\\ \url{https://vista-unitn-uva.github.io}.\label{ftn:website}}
\begin{itemize}[itemsep=1pt] %
\item The introduction of a single visually-grounded dialogue state encoder
jointly trained with the guesser and question generator modules to address
a foundational question of how to integrate visual grounding with dialogue system
components; this yields up to 9\% improvement on task success.

\item The effectiveness of cooperative learning, which yields an additional increase of 8.7\% accuracy, while being easier to train than RL.  %

\item  A first in-depth study to compare cooperative learning  to a state-of-the-art RL system. %
Our study shows that the linguistic skills of the models differ dramatically, despite approaching comparable task success levels. This underlines the importance of linguistic analysis to complement solely numeric evaluation.
\end{itemize}

\section{Related Work}
\label{sec:related}

\paragraph{Task-oriented dialogue systems}  

The conventional architecture of task-oriented dialogue systems includes a pipeline of components, and the task of tracking the dialogue state is typically modelled as a partially-observable Markov decision process~\cite{williams2013state-tracking,young2013pomdp,kim2014inverse}
that operates on a symbolic dialogue state consisting of predefined variables. The use of symbolic representations to characterise the state of the dialogue has some advantages 
(e.g., ease of interfacing with knowledge bases), but it has also some key disadvantages: the variables to be tracked have to be defined in advance and the system needs to be trained on data annotated with explicit state configurations.

Given these limitations, there has been a shift towards neural end-to-end systems
that learn their own representations. Early works focus on non-goal-oriented chatbots~\cite{VinyalsLe:2015,sordoni-EtAl:2015:NAACL-HLT,serban2016building,Li-EtAl:NAACL2016,Li-EtAl:emnlp2016}. 
\newcite{bord:lear17} propose a memory network to adapt an end-to-end system to task-oriented dialogue. %
Recent works combine conventional symbolic with neural
approaches~\cite{will:hybr17,zhao:towa16,Rastogi-Etal-sigdial2018}, but all focus on
language-only dialogue.
We propose a visually grounded task-oriented end-to-end dialogue
system which, while maintaining the crucial aspect of the interaction of
the various modules at play in a conversational agent, grounds them through vision.

\paragraph{Visual dialogue agents}
In recent years, researchers in computer vision have proposed tasks that combine 
visual processing with dialogue interaction. Pertinent datasets created by
\newcite{visdial} and \newcite{guesswhat_game} include \emph{VisDial} and
\emph{GuessWhat?!}, respectively, where 
two participants ask and answer questions about an image. 
While impressive progress has been made in combining vision and language, current
models make simplifications regarding the integration of these two modalities
and their exploitation for task-related actions.  
For example, the models proposed for \textit{VisDial} by
\newcite{visdial} concern an image guessing
game where one agent does not see the target image (thus, no
multimodal understanding) and is required to `imagine' it by asking questions.
The other agent does see the image, but only responds to questions without the need to 
perform additional actions.
 
In {\em GuessWhat?!}, the Questioner agent
 sees an image and asks questions to identify a target 
object in it. The Questioner's role hence involves a complex interaction of 
vision, language, and guessing actions. Most research to date has 
investigated approaches consisting of different models trained independently~\cite{guesswhat_game,stru:end17,zhu:inte17,lee:answ17,shek:askn18,zhan:aski}.
We propose the first multimodal dialogue agent for the \emph{GuessWhat?!}~task 
where all components of the Questioner agent 
are integrated into a joint architecture that has
at its core a \emph{visually-grounded dialogue state encoder}
(cf.~Figure~\ref{fig:model-intro}).

Reinforcement learning for visual dialogue agents was introduced by
\newcite{visdial_rl}  for \textit{VisDial} and by~\newcite{stru:end17} for \textit{GuessWhat?!}. 
Our joint architecture allows us to explore a simpler
solution based on cooperative learning between the agent's internal modules
(see Section~\ref{sec:experiments} for details).

\section{Task and Data}
\label{sec:taskdata}

The {\em GuessWhat?!}~game~\cite{guesswhat_game} is a simplified instance of a
referential communication task where 
two players collaborate to identify a referent---a setting used extensively in
human-human collaborative dialogue \cite{clark1986referring,yule1997referential,zarriess2016pentoref}.

The {\em GuessWhat?!}~dataset\footnote{\url{https://guesswhat.ai/download}} was collected via Amazon Mechanical Turk by \newcite{guesswhat_game}.
The task involves two human participants who see a real-world image, 
taken from the MS-COCO dataset~\cite{lin:micr14}. 
One of the participants (the Oracle) is assigned a target object in the image 
and the other participant (the Questioner) 
has to guess it by asking Yes/No questions to the Oracle. There are no time constraints to play the game. 
Once the Questioner is ready to make a guess, the list of candidate objects is provided and the game is considered successful if the Questioner picks the target object.
The dataset consists of around 155k English dialogues about
approximately 66k different images.~Dialogues contain on average 
5.2 questions-answer pairs.

\section{Models}
\label{sec:models}

We focus on developing an agent who plays the role of the Questioner
in  \emph{GuessWhat?!}.

\subsection{Baseline model}

As a baseline model (BL), we consider our
own implementation of the best performing system put forward by
\citeauthor{guesswhat_game}~\shortcite{guesswhat_game}. It consists
of two independent models: a Question Generator (QGen) and a Guesser. 
For the sake of simplicity, QGen asks a fixed number of questions before the Guesser predicts the target object.

QGen is implemented as an Recurrent Neural Network (RNN) with a transition function handled with Long-Short-Term Memory
(LSTM)~\cite{hochreiter1997long}, on which a probabilistic sequence model is built with a Softmax
classifier. At each time step in the dialogue, the model receives as input the raw image 
and the dialogue history 
and generates the next question one word at a time.
The image is encoded by extracting its VGG-16 features~\cite{simonyan2014very}. 
In our new joint architecture (described below in Section~\ref{sec:GDSE}), 
we use ResNet152~\cite{he2016:resnet}  features instead of VGG, because they tend to yield better performance 
in image classification and are more efficient to compute. For the baseline model it turns out that the original
VGG-16 features lead to better performance (41.8\% accuracy for VGG-16
vs.~37.3\% with ResNet152 features). While we use ResNet152 features
in our models, we keep the original VGG-16 feature
configuration as~\newcite{guesswhat_game}, which constitutes a stronger
baseline.

The Guesser model exploits the annotations in the MS-COCO dataset~\cite{lin:micr14} to represent candidate objects by their object category and their
spatial coordinates. This yields better performance than using raw image features in this case, as reported by \citeauthor{guesswhat_game}~\shortcite{guesswhat_game}. The objects' categories and coordinates are passed through a Multi-Layer Perceptron (MLP) to get an embedding for each object.
The Guesser also takes as input the dialogue history processed by its own dedicated LSTM.
A dot product between the hidden state of the LSTM and each of the object embeddings returns a score for each candidate object. 

The model playing the role of the Oracle is informed about the target object $o_{target}$. Like the Guesser, the Oracle does not have access to the raw image features. It receives as input embeddings of the target object's category, its spatial coordinates, and the current question asked by the Questioner, encoded by a dedicated LSTM. These three embeddings are concatenated and fed
to an MLP that gives an answer (Yes or No).

\subsection{Visually-grounded dialogue state encoder}
\label{sec:GDSE}

In line with the baseline model, our Questioner agent includes two
sub-modules, a QGen and a Guesser. 
As in the baseline, the Guesser guesses after a fixed 
number of questions, which is a parameter tuned on the validation set.
Our agent architecture differs from the baseline model by ~\citeauthor{guesswhat_game}: Rather than operating
independently, the language generation and guessing modules are
connected through a common \emph{grounded dialogue state encoder} (GDSE)
which combines linguistic and visual information as a prior for the
two modules. Given this representation, we will refer to our Questioner agent as GDSE.

As illustrated in Figure~\ref{fig:model-intro}, the encoder receives as input representations of the visual and linguistic
context. The visual representation consists of the second to 
last layer of ResNet152 trained on ImageNet.
The linguistic representation is obtained by an LSTM (LSTM$_e$) which
processes each new question-answer pair in the dialogue.
At each question-answer $QA_t$, the last hidden state of LSTM$_e$ is concatenated with the image features $I$,
 passed through a linear layer and  a \emph{tanh} activation to result in
  the  final layer $h_t$: 
\begin{equation}%
h_t = \tanh \left( W \cdot \left[\textrm{LSTM}_e(qa_{1:t-1});\ I\right] \right)
\end{equation}
where $[\cdot ; \cdot]$ represents concatenation, $I \in \mathbb{R}^{2048 \times 1}$, $\textrm{LSTM}_e \ \in \mathbb{R}^{1024 \times 1}$ and $W \in \mathbb{R}^{512 \times 3072}$ (identical to prior work except for tuning the ResNet-specific parameters).
We refer to this final layer as the \emph{dialogue state}, which is given as input to both QGen and Guesser.

\begin{figure}\centering \hspace*{-0.2cm}
\includegraphics[width=\columnwidth]{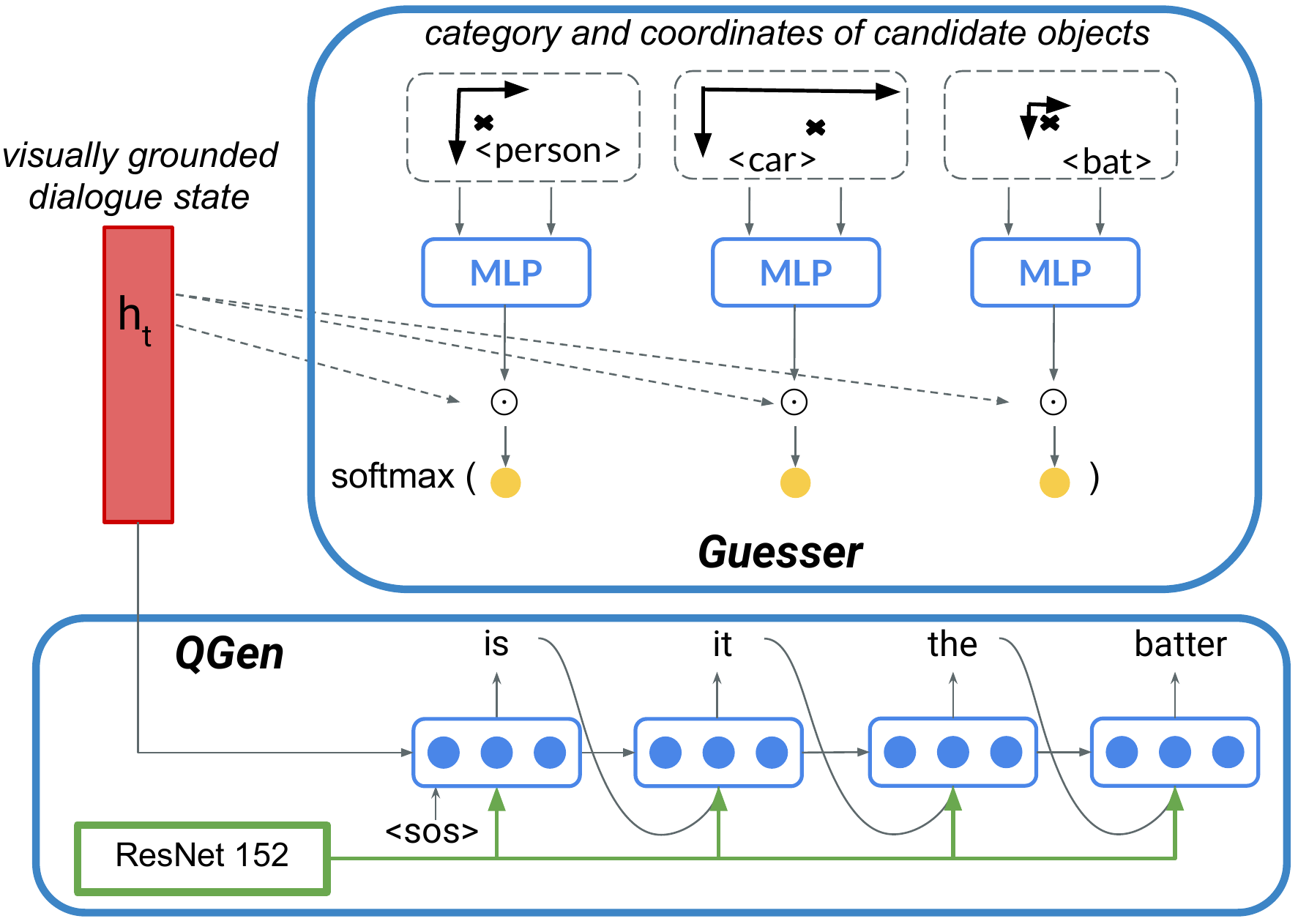}
\caption{Question Generation and Guesser modules.}\label{fig:modules}
\vspace*{-7pt}
\end{figure}

As illustrated in Figure~\ref{fig:modules}, our QGen and Guesser modules are like
  the corresponding modules by~\newcite{guesswhat_game}, except for the 
crucial fact that they receive as input the same grounded dialogue state representation. 
QGen employs an LSTM (LSTM$_q$) to
generate the token sequence for each question conditioned on $h_t$,
which is used to initialise the hidden state of LSTM$_q$.  As input at
every time step, QGen receives a dense embedding of the previously
generated token $w_{i-1}$ and the image features $I$:
\begin{equation}
p(w_i) = p(w_i|w_1, ..., w_{i-1}, h_t, I) \label{eq:tok_prob}
\end{equation}
We optimise QGen by minimising the Negative Log Likelihood (NLL) of the human dialogues and use the Adam optimiser \cite{kingma2014adam}: 

\begin{equation}%
\mathcal{L}_Q = \sum_i - \log p(w_i) \label{eq:qgen_loss}
\end{equation}
Thus, in our architecture the LSTM$_q$ of QGen in combination with the
LSTM$_e$ of the Encoder form a sequence-to-sequence model~\cite{sutskever2014sequence}, conditioned on the visual and linguistic context --- in contrast to the baseline model, where question generation is performed by a single LSTM on its own.

The Guesser consists of an MLP which is evaluated for each candidate object in the image. It takes the dense embedding of the category and the spatial information of the object to establish a representation $r_j \in \mathbb{R}^{512 \times 1}$ for each object. A score is calculated for each object by performing the dot product between the dialogue state $h_t$ and the object representation. Finally, a softmax over the scores results in a probability distribution over the candidate objects:
\begin{equation}%
p(o_j) = \frac{e^{h_t^T \cdot r_j }}{\sum_j e^{h_t^T \cdot r_j }} \label{eq:obj_prob}
\end{equation}
We pick the object with the highest probability 
and the game is successful if $o_{\it guess}\!=\!o_{\it target}$, where 
$o_{guess} = \argmax_j p(o_j) \label{eq:guess_obj}$.
As with QGen, we optimise the Guesser by minimising the NLL and again make use of Adam:
\begin{equation}%
\mathcal{L}_G = - \log p(o_{target}) \label{eq:guess_loss}
\end{equation}

The resulting architecture is fully  differentiable. In addition, the GDSE agent faces a multi-task optimisation problem: While the QGen optimises $\mathcal{L}_Q$ and the Guesser optimises $\mathcal{L}_G$, the parameters of the Encoder ($W$, $\textrm{LSTM}_e$) are optimised via both $\mathcal{L}_Q$ and $\mathcal{L}_G$. 
Hence, both tasks faced by the Questioner agent contribute to the optimisation of the dialogue state $h_t$, and thus to a more effective encoding of the input context.

\section{Learning Approach} 
\label{sec:experiments}

We first introduce the supervised learning approach used to train
both BL and GDSE, then our cooperative learning regime, and
finally the reinforcement learning approach we compare to.

\subsection{Supervised learning}

In the baseline model, the QGen and the Guesser modules are trained autonomously
with supervised learning (SL): QGen is trained to replicate
human questions and, independently, the Guesser is trained to predict the target object.
Our new architecture with a common dialogue state encoder allows us to formulate these
two tasks as a multi-task problem, with two different
losses (Eq.~\ref{eq:qgen_loss} and \ref{eq:guess_loss} in
Section~\ref{sec:GDSE}). %
These two tasks are not equally difficult: While the Guesser has to
learn the probability distribution of the set of possible objects in
the image, QGen needs to fit the distribution of natural language
words.  Thus, QGen has a harder task to optimize and requires more
parameters and training iterations.  We address this issue by making
the learning schedule task-dependent.  We call this setup
\textit{modulo-n} training, where \textit{n} indicates after how many
epochs of QGen training the Guesser is updated together with  QGen.

Using the validation set, we experimented with \textit{n} from 5 to 15 and found that updating
the Guesser every 7 epochs worked best.  With this
optimal configuration, we then train GDSE for 100 epochs (batch size of 1024, Adam, learning rate of 0.0001) and
select the Questioner module best performing on the validation
set (henceforth, GDSE-SL or simply SL).

\subsection{Cooperative learning}

Once the model has been trained with SL, new training data can be
generated by letting the agent play new games. Given an image from the
training set used in the SL phase, we generate a new training instance
by randomly sampling a target object from all objects in the image.
We then let our Questioner agent and the Oracle play the game with
that object as target, and further train the common encoder using
the generated dialogues by backpropagating the error with gradient descent through the
Guesser.  After training the Guesser and the encoder with generated
dialogues, QGen needs to `readapt' to the newly arranged encoder
parameters.  To achieve this, we re-train QGen on the human data with SL, but
using the new encoder states.  Also here, the error is backpropagated with gradient descent
through the common encoder.  

Regarding  \textit{modulo-n}, in this case QGen is updated at every $n^{\rm th}$ epoch,
while the Guesser is updated at all other epochs; %
we experimented with \textit{n} from 3-7 and set it to the optimal value of 5.
The GDSE previously trained with SL is further trained with this
cooperative learning regime for 100 epochs (batch size of 256, Adam, learning rate of 0.0001), and we select the Questioner module performing best on the validation set (henceforth, GDSE-CL or simply CL).

\subsection{Reinforcement learning}\label{sec:rl}
\newcite{stru:end17} proposed the first extension of
BL~\cite{guesswhat_game} with deep reinforcement learning (RL). 
They present an architecture for end-to-end training using an RL policy. %
First,  the Oracle, Guesser, and QGen models are trained independently using supervised learning. Then,
QGen is further trained using a policy gradient.

We use the publicly available code and pre-trained model based on Sampling~\cite{stru:end17},
which resulted in the closest performance to what was reported %
by the authors.\footnote{Their result of 53.3\% accuracy
  published in \newcite{stru:end17} is obsolete, as stated on their
  GitHub page~(\url{https://github.com/GuessWhatGame/guesswhat}) where
  they report 56.5\% for sampling and 58.4\% for greedy search. By
  running their code, we could only replicate their results with
  sampling, obtaining 56\%, while greedy and beam search resulted in similar or
  worse performance. Our analysis showed that greedy and beam search have the additional 
  disadvantage of learning a smaller vocabulary.} 
This is the RL model we use throughout the rest of the paper.

\subsection{Experimental details}

We use the same train (70\%), validation (15\%), and
test (15\%) splits as~\newcite{guesswhat_game}. The test set
contains new images not seen during training. 
We use two experimental setups for the number of questions 
to be asked by the question generator, motivated by prior work: 5 questions (5Q) following~\newcite{guesswhat_game}, and 8 questions (8Q) as in \newcite{stru:end17}.
As noted in Section~\ref{sec:taskdata}, on average, there are 5.2 questions per dialogue in the \textit{GuessWhat?!} data set.

For evaluation, we report task success in terms of accuracy~\cite{stru:end17}.
To neutralize the effect of random sampling in training CL, we trained the model 3 times. RL is 
tested 3 times with sampling. We report means and standard deviation (for some tables these are provided in the supplementary material; see footnote \ref{ftn:website}).

\section{Results}
\label{sec:results}

Table~\ref{tab:accuracy} reports the results for all models. There are several take-aways.

\paragraph{Grounded joint architecture} 
First of all, our visually-grounded dialogue state encoder
is effective. GDSE-SL outperforms the baseline by~\newcite{guesswhat_game} 
significantly in both setups (absolute accuracy improvements of 6.6\% and 9\%).
To evaluate the impact of the multi-task learning aspect, we did an
ablation study and used the encoder-decoder architecture to train
the QGen and Guesser modules independently. With such a decoupled
training we obtain lower results: 44\% and 43.7\% accuracy for 5Q and 8Q,
respectively. Hence, the multi-task component brings an increase
  of up to 6\% over the baseline.\footnote{While
 \citeauthor{guesswhat_game}~\shortcite{guesswhat_game} originally
  report an accuracy of 46.8\%, this result was later revised to
  40.8\%, as clarified on their GitHub page. Our own implementation of
  the baseline system achieves an accuracy of 41.2\%.} 

\begin{table}\centering 
\resizebox{0.8\columnwidth}{!}{ %
\begin{tabular}{llll}\toprule
Model & 5Q & 8Q \\\midrule
Baseline & 41.2 & 40.7 \\
GDSE-SL & 47.8& 49.7  \\
GDSE-CL  & 53.7 ($\pm$.83) & 58.4 ($\pm$.12) \\  %
RL  & 56.2 ($\pm$.24) & 56.3 ($\pm$.05) \\
\bottomrule
\end{tabular}
} %
\caption{Test set accuracy for each model (for setups with 5 and 8 questions). GDSE-SL is our grounded supervised learning system, GDSE-CL the cooperative learning setup, and RL the results we obtain with the reinforcement learning system by Strub et al. (2017).}\label{tab:accuracy}
\end{table}

\paragraph{Cooperative learning and RL} 
The  introduction of the cooperative learning approach results in a clear improvement over GDSE-SL: +8.7\% (8Q: from 49.7 to 58.4) and +5.9\% (with 5Q).
Despite its simplicity, our GDSE-CL model achieves a task success rate which is comparable to
RL: %
 In the 8Q setup, GDSE-CL reaches an average accuracy of 58.4 versus 56.3 for RL, 
 giving CL a slight edge in this setup (+2.1\%), while in the 5Q setup RL is slightly better (+2.5\%). 
Overall, the accuracy of the CL and RL models is close.  The interesting question is how the
linguistic skills and strategy of these two models differ, to which we turn in the next section.

We compared to~\newcite{stru:end17}, but RL has
also been put forward by~\newcite{zhan:aski}, who
report 60.7\% accuracy (5Q). This result is close
to our highest GDSE-CL result (60.8 $\pm$0.51, when
optimized for 10Q).\footnote{Since our aim is to compare to the best setup 
for BL (5Q) and RL (8Q), we do not report our results with 10Q in Table~\ref{tab:accuracy}.} Their RL system integrates several partial reward 
functions to increase coherence, which
is an interesting aspect. Yet their code is not publicly
available. We leave the comparison to~\newcite{zhan:aski} and adding RL to GDSE to future work.

\section{Analysis}
\label{sec:analysis}

In this section, we present a range of analyses that aim to shed light
on the performance of the models. They are carried out on the test set data 
 using the 8Q setting, which yields better results than the 5Q setting for the GDSE models and RL.
Given that there is only a small difference in accuracy for the baseline with 5Q and 8Q, 
for comparability we analyse dialogues with 8Q also for BL.

\subsection{Quantitative analysis of linguistic output}

We analyse the language produced by the Questioner agent with respect to three factors:
(1) lexical diversity, measured as type/token ratio over all games, 
(2) question diversity, measured as the percentage of unique questions over all games, and 
(3) the number of games with questions repeated verbatim.
We compute these factors on the test set for the models and
for the human data (H). 

As shown in Table~\ref{tab:ling-output},
the linguistic output of SL \& CL is closer to the language used by humans: 
Our agent is able to produce a much richer and less repetitive output 
than both BL and RL. %
In particular, it learns to use a more diverse vocabulary, 
generates more unique questions, 
and repeats questions within the same dialogue at a much lower rate %
than the baseline and RL: 93.5\% of the games played by BL contain at least
one verbatim question repetition, for RL  this happens in 96.47\% of the cases, whereas for SL and CL this is for only 55.8\% and 52.19\% of the games, respectively.

\begin{table}[ht!]\centering
\resizebox{0.95\columnwidth}{!}{
\begin{tabular}{llll} \toprule
 & \parbox{2cm}{Lexical\\ diversity} & \parbox{2cm}{Question\\ diversity} & \parbox{2.3cm}{\% Games with\\ repeated Q's} \\ \midrule
BL & 0.030 & \ \ 1.60 & 93.50 \\
SL & 0.101 & 13.61 & 55.80 \\
CL & 0.115 ($\pm$.02) & 14.15 ($\pm$3.0) & 52.19 ($\pm$4.7) \\
RL & 0.073 ($\pm$.00) & \ \ 1.04 ($\pm$.03) &  96.47 ($\pm$.04) \\
H & 0.731 & 47.89 & --- \\ \bottomrule
\end{tabular}
} %
\caption{Statistics of the linguistic output of all models with the 8Q setting and of humans (H) in all test games.
}\label{tab:ling-output}
\vspace*{-10pt}
\end{table}

\subsection{Dialogue strategy}

\begin{figure*}
\resizebox{\textwidth}{!}{
\begin{tabular}{@{}l@{}c|c|c@{}}%
\begin{minipage}[c]{4.3cm}
\includegraphics[width=4.3cm]{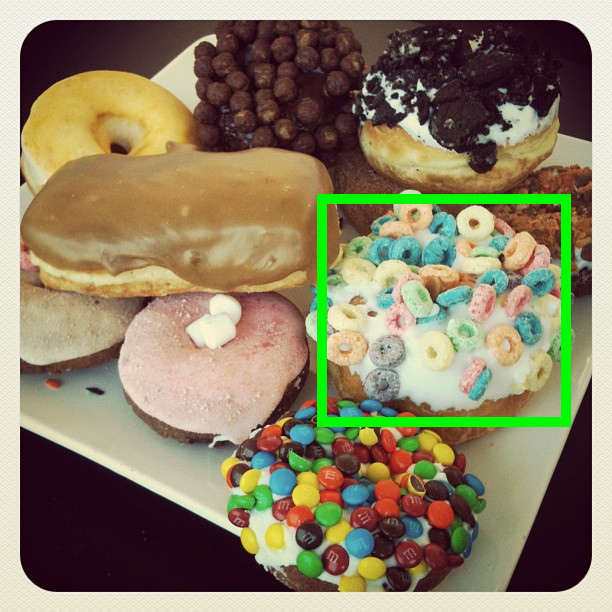}  
\end{minipage} & 
\begin{tabular}{l@{\ }r}
\multicolumn{2}{l@{\ }}{\textbf{Humans} \hfill [{\em success}]} \\
1. does it have cereal on it? & no\\
2. does it have pink frosting? & no\\
3. does it have cookies? & no\\
4. is it a donut? & yes\\
5. does it have maple frosting? & no\\
6. is there candy? & yes\\
7. is it the doughnut with M\&Ms? & no\\
8. is it dark brown? & no\\
\end{tabular} &
\begin{tabular}{@{}l@{\ }r}
\multicolumn{2}{@{}l@{\ }}{\textbf{GDSE-CL} \hfill [{\em success}]} \\
1. is it a donut? & yes\\
2. is it on the left? & no\\
3. is it on the right? & yes\\
4. is it the whole donut? & yes\\
5. does it have pink icing? & no\\
6. is it touching the donut with the sprinkles? & yes\\
7. does it have chocolate icing? & no\\
8. is it the third donut? & yes
\end{tabular} & 
\begin{tabular}{@{}l@{\ }r@{\ }}
\multicolumn{2}{@{}l@{\ }}{\textbf{RL} \hfill [{\em failure}]} \\
1. is it food? & yes\\
2. is it a donut? & yes\\
3. is it in left? & no\\
4. is it in top? & yes\\
5. is it in top? & yes\\
6. is it in top? & yes\\
7. is it in top? & yes\\
8. is it top? & yes
\end{tabular} \\ \bottomrule
\end{tabular}
} %
\caption{Game example where GDSE-CL succeeds and RL fails at guessing the target object (green box).}
\label{fig:ex}
\end{figure*}

To further understand the variety of questions asked by the agents, we classify
questions into different types.
We distinguish between questions that aim at getting the category of the target object ({\sc entity} questions, e.g., {\em `is it a vehicle?'})
and questions about properties of the queried objects ({\sc attribute} questions, 
e.g.,  {\em `is it square?'} or {\em `are they standing?'}). 
Within {\sc attribute} questions, 
we make a distinction between color, shape, size, texture, location, and action questions. 
Within {\sc entity} questions, we distinguish questions whose focus is an
object category or a super-category  (see the supplementary material for example questions). %
The classification is done by manually
extracting keywords for each question type from the human dialogues, and then
applying an automatic heuristic that assigns a class to a question 
given the presence of the relevant keywords.\footnote{A question may be tagged with several attribute classes if keywords of different types are present. E.g., {\em ``Is it the white one on the left?''}~is classified as both {\sc color} and {\sc location}.} 
This procedure allows us to classify 91.41\% of the questions asked by humans.
The coverage is higher for the questions asked by the models: 98.88\% (BL), 94.72\%
(SL), 94.11\% (CL) and 99.51 \% (RL).\footnote{In the supplementary material we
  provide details on the question classification procedure: the lists of keywords by class, the procedure used to obtain these lists, as well as the pseudo-code of the heuristics used to classify the
  questions.}  

The statistics are shown in Table~\ref{tab:question-types}.
We use Kullback-Leibler (KL) divergence to measure how the output of each model differs from the human distribution of fine-grained question classes. The baseline's output has the highest degree of divergence: For instance, the BL model does never ask any {\sc shape} or {\sc texture} questions, and hardly any {\sc size} questions. The output of the RL model also differs substantially from the human dialogues: It asks a very large number of {\sc location} questions (74.8\% vs.~40\% for humans). Our model, in contrast, generates question types that resemble the human distribution more closely.

\begin{table}[h!]\centering
\resizebox{0.87\columnwidth}{!}{%
\begin{tabular}{@{}l@{\hspace*{-.28cm}}c@{\ \ }ccccc@{}}\toprule
 Question type  & BL & SL  & CL & RL & H\\\midrule
\textbf{\textsc{entity}} & \textbf{49.00}& \textbf{48.07} &  \textbf{46.51}& \textbf{23.99}& \textbf{38.11}  \\
 \sc super-cat  & 19.6 & 12.38 
                                        & 12.58 & 14.00 & 14.51\\
 \sc object    & 29.4 & 35.70  & 33.92 & 9.99 & 23.61\\
\textbf{\textsc{attribute}}  & \textbf{49.88} & \textbf{46.64} &
                                                                   \textbf{47.60} & \textbf{75.52}& \textbf{53.29}\\
 \sc color  & 2.75 & 13.00  & 12.51 & 0.12 & 15.50 \\
 \sc shape & 0.00 & 0.01 &  0.02 & 0.003 & 0.30  \\ 
 \sc size & 0.02 & 0.33 & 0.39 & 0.024 & 1.38  \\
 \sc texture & 0.00 & 0.13 &  0.15 & 0.013 & 0.89  \\
 \sc location & 47.25 & 37.09 & 38.54
                                        & 74.80 & 40.00  \\
 \sc action  & 1.34 & 7.97 &  7.60 & 0.66 & 7.59  \\
\textbf{Not classified} & \textbf{1.12} & \textbf{5.28} &
                                                             \textbf{5.90}
                                        & \textbf{0.49} & \textbf{8.60}\\ \midrule %
\multicolumn{1}{@{}l@{\ }}{KL (wrt human)} &  0.953 &  0.042 &  0.038 & 0.396 &  0.0 \\ \bottomrule %
\end{tabular} 
} %
\caption{Percentage of questions per question type in all the test set games
 played by humans (H) and the models with the 8Q setting, and 
 KL divergence from human distribution of fine-grained question types. 
} \label{tab:question-types}
\end{table}

 \begin{figure*}[h!]
\resizebox{1.02\textwidth}{!}{
  \subfloat[a][Lexical diversity]{
  \hspace*{-.6cm}    \includegraphics[height=3cm]{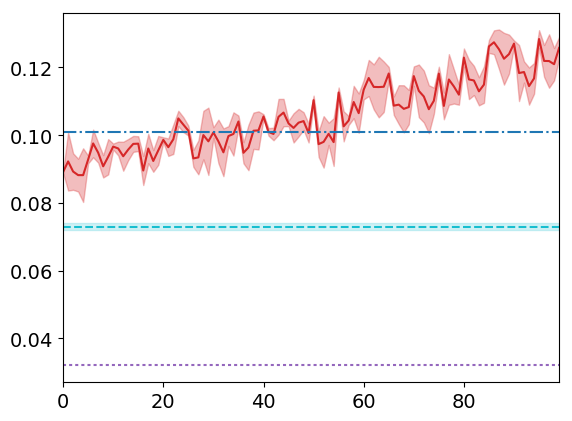}}
  \subfloat[b][Question diversity]{
  \hspace*{-.25cm}  \includegraphics[height=3cm]{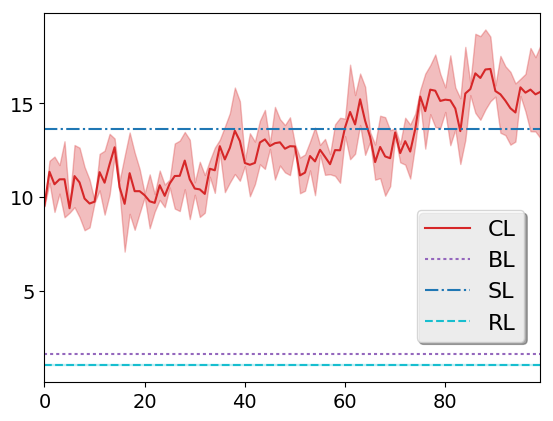}}
  \subfloat[c][\% Games w/ repeated Q's]{
  \hspace*{-.2cm}     \includegraphics[height=3cm]{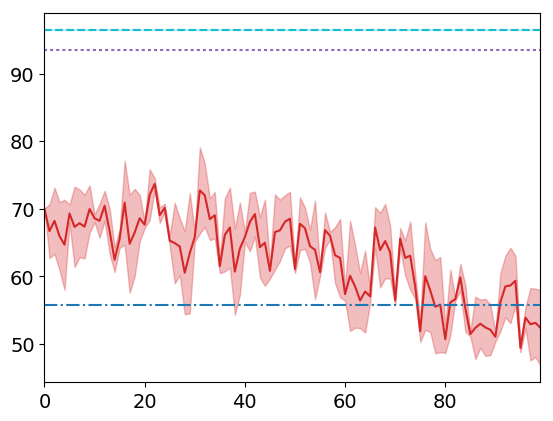}}
  \subfloat[d][KL-distance from human]{
   \hspace*{-.18cm}\includegraphics[height=3cm]{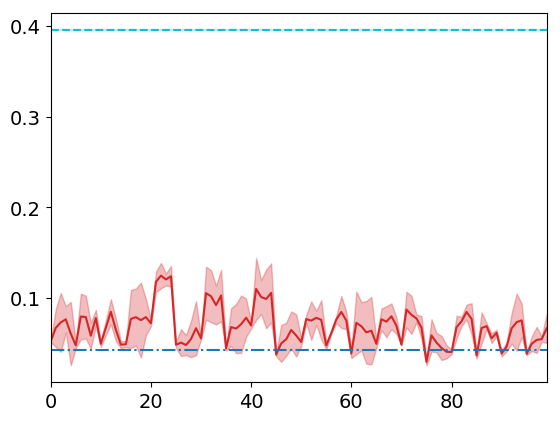}}
} %
\caption{Evolution of linguistic factors over 100 training epochs for our GDSE-CL model.
   Note: lexical and question diversity of the human data fall outside the range in (a) / (b). The same is the case with KL for BL in (d).
    }
    \label{fig:cl}
\end{figure*}

We also analyse  the structure of
the dialogues in terms of the sequences of question types asked.  As
expected, both humans and models  almost always start with an {\sc
  entity} question (around 97\% for BL, SL and CL, 98.7\% for RL, and 78.48\% for
humans), in particular a {\sc super-category} (around 70\% for BL, SL
and CL, 84\% for RL, and 52.32\% for humans). In some cases, humans start by asking
questions directly about an attribute that may easily distinguish an object
from others, while this is very uncommon for models. 
Figure~\ref{fig:ex} shows an example: The human dialogue begins with an {\sc attribute} question
({\em `does it have cereal on it?'}), which in this case is not very effective and leads to
a change in strategy at turn 4.
The CL model starts by asking an {\sc object} question ({\em `is it a donut?'})
while the RL model begins with a more generic {\sc super-category} question ({\em `is it food?'}).

We check how the answer to a given question type affects the type of the follow-up question.
In principle, we expect to find that question types that are answered positively 
will be followed by more specific questions. This is indeed what we observe
in the human dialogues, as shown in Table~\ref{tab:q-change}. 
For example, when a {\sc super-category} question
is answered positively, humans follow up with an {\sc object} or {\sc attribute} question 89.56\%  of the time.
This trend is mirrored by all models. 

\begin{table}[h]
\resizebox{\columnwidth}{!}{
\begin{tabular}{@{}lccccc@{}}\toprule
Question type shift  & BL & SL  & CL & RL & H\\\midrule
{\sc super-cat} $\rightarrow$ {\sc obj}/{\sc att}
    & 89.05 & 92.61 & 89.75 & 95.63 & 89.56\\
{\sc object} $\rightarrow$ {\sc attribute}
	& 67.87 & 60.92 & 65.06 & 99.46 & 88.70\\\bottomrule
\end{tabular} 
}
\caption{Proportion of question type shift vs.~no type shift in 
consecutive questions $Q_t \rightarrow Q_{t+1}$ where $Q_t$ has received a Yes answer.}\label{tab:q-change}
\end{table}

\noindent
Overall, the models also learn the strategy to move from an {\sc object} to an {\sc attribute} question when an {\sc object} question receives a Yes answer. The BL, SL, and CL models do this to a lesser extent than humans, while the RL model systematically transitions to attributes (in 99.46\% of cases),  using mostly {\sc location} questions as pointed out above. For example (Figure~\ref{fig:ex}), after receiving an affirmative answer to the {\sc object} question `is it a donut?' both CL and RL shift to a {\sc location} question. Once location is established, CL moves on to other attributes while RL keeps asking the same {\sc location} question, which leads to failure. Further illustrative examples are given in the supplementary material.

\subsection{Analysis of the CL learning process}

In order to better understand the effect of the cooperative learning
regime, we trace the evolution of linguistic factors identified above over the 
CL epochs. 
As illustrated in Figure~\ref{fig:cl} (a) and (b), through the epochs the CL model learns
to use a richer vocabulary and more diverse questions, moving away from the levels achieved by BL and RL, overpassing SL and moving toward humans. %

The CL model progressively produces fewer repeated questions within a dialogue, improving over SL in the last few epochs, cf.\ Figure~\ref{fig:cl} (c).  %
Finally, (d) illustrates 
the effect of modulo-$n$ training: As the model is trained on generated dialogues, its linguistic output drifts away from the human distribution of question types; every 5$^{th}$ epoch QGen is trained via supervision, which brings the model's behaviour closer back to 
human linguistic style and helps decrease the drift. %

\section{Conclusion}
\label{sec:conclusion}
We present a new visually-grounded joint Questioner agent for goal-oriented dialogue. %
First, we show that our architecture
archives 6--9\% accuracy improvements over
the \emph{GuessWhat?!} baseline system~\cite{guesswhat_game}.  
This way, we address a foundational limitation of previous approaches that model 
guessing and questioning separately.

Second, our joint architecture allows us to propose a two-phase cooperative
learning approach (CL), which further improves accuracy. It results in our
overall best model and reaches state-of-the-art results (cf.\ Section~\ref{sec:results}).
We compare CL to the
system proposed by~\newcite{stru:end17} which extends the baseline with reinforcement learning (RL).
We find that the two approaches (CL and RL) achieve overall relatively
similar task success rates. However, evaluating on task success is only one
side of the coin.
Finally and most importantly, we propose to pursue an in-depth analysis 
of the quality of the dialogues by visual conversational agents, which
 is an aspect often neglected in the literature. We analyze the linguistic
output of the two models across three factors (lexical diversity, question diversity, and
repetitions) and find them to differ substantially.  The
CL model uses a richer vocabulary and inventory of questions, and produces fewer repeated questions than
RL. In contrast, RL highly relies on asking location questions, which
might be explained by a higher reliance on spatial and object-type
information explicitly given to the Guesser and Oracle models.
Limiting rewards to task success or other rewards not connected to the language proficiency does not stimulate the model to learn 
rich linguistic skills, since a reduced vocabulary and simple linguistic structures 
may be an efficient strategy to succeed at the game. 

Overall, the presence of repeated questions remains an important
weakness of all models, resulting in unnatural dialogues. This shows that there is still a considerable gap to human-like
conversational agents. Looking beyond task success can provide a good basis for
extensions of current architectures, e.g.,~\newcite{shek:askn18} add a decision-making component that decides when to stop asking 
questions which results in less repetitive  and more human-like dialogues. Our joint architecture could
easily be extended with such a component.

\section*{Acknowledgements}
The work carried out by the Amsterdam team was partially funded by the Netherlands Organisation for Scientific Research (NWO) under VIDI grant nr.~276-89-008, {\em Asymmetry in Conversation}.  We thank the University of Trento for generously funding a research visit by Barbara Plank to CIMeC that led to part of the work presented in this paper. In addition, we kindly acknowledge the support of NVIDIA
 Corporation with the donation to the University of Trento of the GPUs used in our research.

\bibliography{raffa,raq,tim}
\bibliographystyle{acl_natbib}

\end{document}